\documentclass{article}
\usepackage{arxiv}

\usepackage[utf8]{inputenc} 
\usepackage[T1]{fontenc}    
\usepackage{hyperref}       
\usepackage{url}            
\usepackage{booktabs}       
\usepackage{amsfonts}       
\usepackage{nicefrac}       
\usepackage{microtype}      
\usepackage{lipsum}
\usepackage{graphicx}
\usepackage{float}
\usepackage{amsmath}
\usepackage{amssymb}
\usepackage{mathtools}
\usepackage{amsthm}
\usepackage{caption}
\usepackage{titlesec}
\usepackage{setspace}
\newtheorem{assumption}{Assumption}

\titlespacing*{\section}{8pt}{12pt}{8pt}
\titlespacing*{\subsection}{8pt}{8pt}{8pt}
\titlespacing*{\subsubsection}{8pt}{8pt}{8pt}

\usepackage{authblk}
\usepackage{savesym}
\usepackage{moreverb}
\savesymbol{bigtimes}
\usepackage{mathabx}
\restoresymbol{TXF}{bigtimes}
\usepackage{makecell}
\usepackage{subcaption}
\usepackage{fancyvrb}
\usepackage{multirow}
\usepackage{booktabs} 
\usepackage{enumitem}
\usepackage{tabularx}
\usepackage[authoryear, round]{natbib}
\usepackage{lmodern}
\usepackage{anyfontsize}
\RequirePackage{fancyhdr}
\usepackage[dvipsnames]{xcolor}
\RequirePackage{algorithm}
\RequirePackage{algorithmicx}
\usepackage{algcompatible}

\RequirePackage{eso-pic} 
\RequirePackage{forloop}
\RequirePackage{url}

\newcommand{\INPUT}[1]{\STATEx \textbf{input:} #1 }
\newcommand{\OUTPUT}[1]{\STATEx \textbf{output:} #1}

\usepackage[most]{tcolorbox}

\newcounter{boxcount} 
\newtcolorbox{examplebox}[1]{ 
  colback=gray!10,
  colframe=gray!50,
  boxrule=0.5pt,
  arc=2pt,
  left=6pt,
  right=6pt,
  top=6pt,
  bottom=6pt,
  code={\refstepcounter{boxcount}}, 
  title={Box \theboxcount: #1},     
  fonttitle=\bfseries,
  coltitle=black,
  attach title to upper
}

\title{Structured Hybrid Mechanistic Models for Robust Estimation of Time-Dependent Intervention Outcomes}

\author[1, $\dag$]{Tomer Meir}
\author[1]{Ori Linial}
\author[1]{Danny Eytan}
\author[2]{Uri Shalit}

\affil[1]{Technion - Israel Institute of Technology}
\affil[2]{Tel Aviv University}
\affil[$\dag$]{Corresponding Author: tomer.me@campus.technion.ac.il}

\begin{document}
\maketitle
\begin{abstract}
Estimating intervention effects in dynamical systems is crucial for outcome optimization. In medicine, such interventions arise in physiological regulation (e.g., cardiovascular system under fluid administration) and pharmacokinetics, among others. Propofol administration is an anesthetic intervention, where the challenge is to estimate the optimal dose required to achieve a target brain concentration for anesthesia, given patient characteristics, while avoiding under- or over-dosing. The pharmacokinetic state is characterized by drug concentrations across tissues, and its dynamics are governed by prior states, patient covariates, drug clearance, and drug administration.
While data-driven models can capture complex dynamics, they often fail in out-of-distribution (OOD) regimes. Mechanistic models on the other hand are typically robust, but might be oversimplified. We propose a hybrid mechanistic-data-driven approach to estimate time-dependent intervention outcomes. 
Our approach decomposes the dynamical system's transition operator into parametric and nonparametric components, further distinguishing between intervention-related and unrelated dynamics. This structure leverages mechanistic anchors while learning residual patterns from data. For scenarios where mechanistic parameters are unknown, we introduce a two-stage procedure: first, pre-training an encoder on simulated data, and subsequently learning corrections from observed data. 
Two regimes with incomplete mechanistic knowledge are considered: periodic pendulum and Propofol bolus injections. Results demonstrate that our hybrid approach outperforms purely data-driven and mechanistic approaches, particularly OOD. This work highlights the potential of hybrid mechanistic-data-driven models for robust intervention optimization in complex, real-world dynamical systems.
\end{abstract}


\begin{figure*}[!ht]
    \centering
    \includegraphics[width=\textwidth, trim=0 3.1cm 0 0, clip]{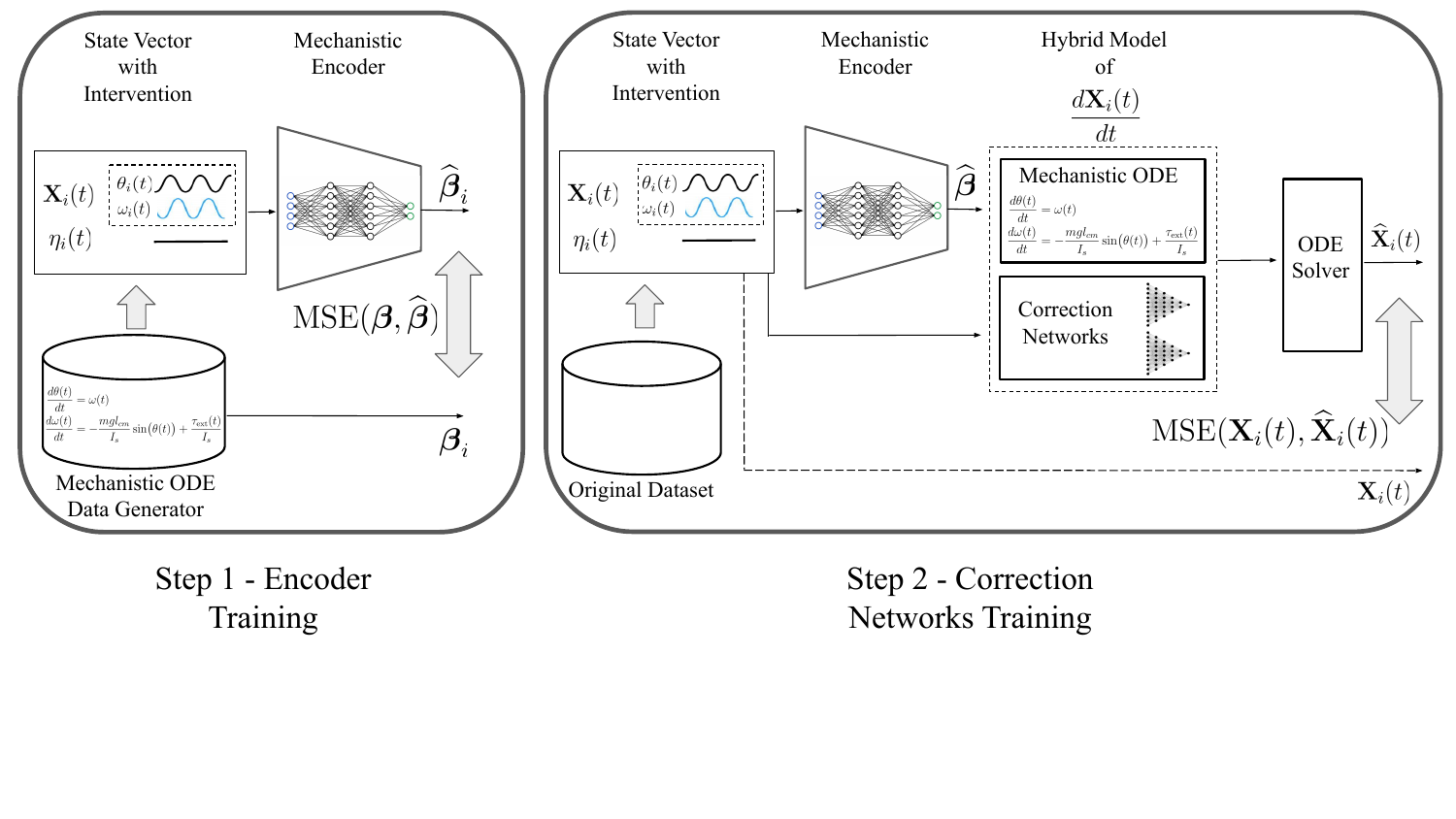}
    \caption[Hybrid model - Training Process]{A hybrid mechanistic-data-driven model with a two-step training process. In Step~1 (left), we generate synthetic training data from the known mechanistic model, which provides ground-truth parameter labels. This labeled data is used to train an encoder that maps the state trajectory and interventions to the parameter vector, using a mean squared error (MSE) loss between true and predicted parameters. In Step~2 (right), we train the correction networks using the original dataset while keeping the encoder network fixed, optimizing an MSE reconstruction loss between the observed and reconstructed signals.}
    \label{fig:training-process}
\end{figure*}

\section{Introduction}

Predicting the trajectory of a dynamical system under varying interventions is a prerequisite for optimal decision-making. In medicine, interventions in dynamical systems are common and include, for example, fluid administration in cardiovascular systems and joint interventions in musculoskeletal systems, among others.
For example, determining the optimal dose for a patient, a critical task in anesthesia, diabetes management and more. In such cases, the system's response is dictated by its pharmacokinetic (PK) state, while the underlying dynamics are driven by drug administration, clearance rates, and patient covariates.
For example, consider Propofol administration, a standard intervention in anesthesia: the challenge lies in estimating the optimal infusion dose schedule to achieve the target concentration in the brain required for anesthesia given patient's characteristics, without incurring the adverse effects of Propofol under- or over- dose.

A central challenge in causal modeling of dynamical systems is that interventions do not merely shift the distribution of a single outcome at a fixed time point, but rather modify the underlying transition operator that maps one state to the next.
Estimating trajectories under interventions therefore requires characterizing the transition operator and the full path of potential states under different treatments \citep{lim_rmsn_2018}. In such settings, even minor structural errors in the transition operator can accumulate over time, causing predicted trajectories to diverge substantially at later horizons. Yet in many biomedical applications the governing dynamics are only partially known, and are based on small sample size -- we do not have anything close to an accurate, complete dynamical-system characterization of the human body.
In this work we develop a hybrid mechanistic-data-driven approach for estimating the effects of interventions in dynamical systems. We focus on settings where the system state is observed, and where interventions are time-dependent but known \emph{a priori} (i.e., not adaptive). An example of such a scenario is choosing a medication schedule for a patient based on their baseline covariates and state at time $t_0$ so as to reach a desired clinical state post treatment.

Modeling the transition operator in these settings raises a familiar tradeoff \citep{karpatne_theory_2017, baker_mechanistic_2018}. On one hand, mechanistic models (often specified as ordinary differential equations) encode prior scientific knowledge through parametric structure; data is then used primarily to estimate a very small set of parameters describing the overall system. These models can be interpretable and stable, but in practice they are frequently misspecified due to missing mechanisms, unmodeled heterogeneity, or inaccurate parameterization. On the other hand, fully nonparametric models (e.g., neural networks) can represent complex dynamics with minimal prior assumptions, but their learned relationships may be difficult to interpret and can be brittle when deployed outside the training distribution. This brittleness is particularly problematic for causal questions: prediction under interventions is inherently performed under a distribution shift, and models that rely on statistical associations rather than causal structure may fail in precisely the regimes of interest \citep{scholkopf_toward_2021}. Our goal is to retain the stability conferred by mechanistic structure while using data-driven components to absorb model mismatch.

This tension has motivated hybrid approaches that combine the strengths of both paradigms, leveraging the stability  of mechanistic models together with the flexibility of nonparametric components \citep{baker_mechanistic_2018, raissi_pinn_2019, rackauckas_universal_2020, miller_breimans_2021}. 
We demonstrate the utility of the hybrid approach through settings where mechanistic knowledge can be expressed through ordinary differential equations (ODEs) and combine this mechanistic structure with nonparametric components learned from data. Specifically we choose to use Neural ODE models \citep{chen_neural_2018} for this purpose, due to their flexibility and wide use. Our framework models the transition operator as a sum of parametric and nonparametric contributions, enabling the correction of mechanistic misspecification while retaining stability. Notably, we do not attempt to identify the ``missing'' dynamics parametrically, as in approaches such as SINDy or CONFIDE \citep{brunton_sindy_2016, linial_confide_2024}. Instead, we intentionally leave nonparametric components to absorb model mismatch while keeping the known mechanistic core intact.
Crucially, we explicitly distinguish between intervention-dependent and intervention-independent corrections. This structural decomposition allows the model to disentangle natural dynamic drift from treatment effects, yielding improved modeling of counterfactual intervention outcomes. A formal presentation of the general dynamical-system framework and our hybrid modeling approach is provided in Section~\ref{sec:neural-ode-setup}.

To demonstrate the versatility of our approach, we examine two case studies involving incomplete mechanistic models. First, we analyze a pendulum case study to evaluate the model's ability to recover dynamics under parametric uncertainty and distributional shifts in intervention magnitude. We propose a training procedure to handle unknown physical parameters.

Second, we address anesthesia dosage control, a PK challenge requiring sequential bolus injections to maintain patient-specific target concentrations. Unlike the pendulum, the patient covariates here are known, but the mechanistic model is structurally deficient \citep{enlund_tci_2008}. We evaluate robustness against OOD shifts in patient parameters, which also affects intervention regimes.

\subsection{Related Work}

Mechanistic ODEs are widely used in physiology, including, for example, PK \citep{eleveld_pharmacokinetic_2018, schnider_influence_1998} or cardiovascular \citep{tannenbaum_icvs_2023}. While interpretable and stable, they oversimplify complex biological interactions, motivating hybrid approaches that retain mechanistic stability while capturing unmodeled dynamics \citep{grigorian_hybrid_2024}. Indeed, hybrid mechanistic-data-driven modeling has gained increasing attention in scientific machine learning. Such models, also referred to as grey-box models, SciML, or physics-informed neural networks (PINNs), integrate mechanistic structure with trainable components to improve robustness \citep{rueden_informed_2023}. Applications span systems biology \citep{lan_shallow_2025, noordijk_rise_2024}, engineering \citep{willard_integrating_2022}, agronomy \citep{maestrini_mixing_2022}, and physics \citep{karniadakis_physics_2021}. 

Semiparametric ODE models \citep{xue_parameter_2019} use nonparametric components for state-smoothing and parameter estimation, however, they assume the underlying mechanistic structure is correct. In contrast, our framework treats the mechanistic model as a potentially incomplete prior, utilizing neural components to learn and compensate for missing or misspecified dynamical terms.

Neural ordinary differential equations (Neural ODEs) \citep{chen_neural_2018} provide a natural framework for learning continuous-time dynamics and have been extended to incorporate mechanistic priors  \citep{rackauckas_universal_2020}. Hybrid dynamical systems approaches such as latent ODEs \citep{rubanova_latent_2019} or neural controlled differential equations \citep{kidger_neuralcde_2020} introduce flexible data-driven dynamics, yet they do not incorporate explicit mechanistic models or exploit known scientific structure. Additional work handles known unknowns \citep{linial_generative_2021}. In contrast to these approaches, we focus on counterfactual questions by introducing a structural distinction between intervention-dependent and intervention-independent components. Moreover, we propose a principled method for estimating the parameters of the mechanistic prior when they are unavailable and demonstrate how such priors facilitate generalization in OOD regimes inherent to causal inference.

Purely data-driven models achieve strong predictive accuracy in time-series forecasting \citep{lim_rmsn_2018}, yet they typically rely on statistical associations rather than causal or mechanistic relations. This limits their performance in counterfactual or OOD settings common in medicine \citep{scholkopf_toward_2021}. Related work in model-based reinforcement learning and differentiable simulation \citep{chua_pets_2018, deisenroth_pilco_2011} similarly focuses on optimal decision making under partially learned dynamics. However, these approaches often prioritize policy optimization over the causal validity of the underlying transition operator. In clinical settings, we argue that optimal decision making must be grounded in physical or physiological laws to ensure safety and robustness in OOD regimes - a fundamental requirement for reliable causal inference.

From a causal inference perspective, \citet{lok_statistical_2008} established the theoretical basis for potential outcomes in continuous time. Subsequent methods have addressed treatment effect estimation in both unconfounded \citep{soleimani_treatment_2017, schulam_reliable_2017} and confounded settings \citep{bica_time_2020}. However, these approaches typically model dynamics via flexible, nonparametric approaches and do not exploit the rich, albeit imperfect, mechanistic knowledge available in biomedical domains. Our work bridges this gap by anchoring the transition operator to a mechanistic core while utilizing nonparametric components to correct for misspecification, specifically tailored for intervention effect estimation.

\subsection{Main Contributions}
This work makes the following contributions:
\begin{itemize}[leftmargin=*, topsep=3pt, noitemsep]
    \item \textbf{Structural decomposition for counterfactual dynamics:} We introduce a hybrid architecture that partitions the transition operator into parametric versus nonparametric, and intervention-dependent versus independent components. This explicit inductive bias facilitates robust counterfactual prediction compared to purely data-driven Neural ODEs.
    \item \textbf{Two-stage inference for latent parameters:} We propose a framework to handle unobserved physical parameters via simulation-based encoder pre-training. This amortized inference provides a grounded initialization for the mechanistic core, preventing the optimization pathology where hybrid models bypass physical constraints to rely solely on data-driven components.
    \item \textbf{Robustness to interventional distribution shift:} We demonstrate that anchoring nonparametric corrections to a mechanistic core significantly improves generalization in OOD regimes. Experiments on pendulum dynamics and Propofol PK show superior stability in counterfactual scenarios involving unseen intervention magnitudes and patient covariates.
\end{itemize}


\section{Setup}
\label{sec:neural-ode-setup}

We consider a dynamical system whose state evolution is governed by the following ODE model:    
\begin{align}
\label{eq:xue-general-form}
    \frac{d\textbf{X}(t)}{dt} &= \widetilde{\mathbf{F}}\{t, \textbf{X}(t), \boldsymbol{\beta}, \boldsymbol{\eta}(t)\}, 
    \qquad \forall t \in [t_0, t_{\max}], \nonumber \\
    \textbf{X}(t_0) &= \textbf{X}_0.
\end{align}
Here, $t \in [t_0, t_{\max}]$ denotes continuous time, $\textbf{X}(t) \in \mathcal{X} \subseteq \mathbb{R}^{k}$ is the $k$-dimensional state vector representing the system at time $t$, and $\textbf{X}(t_0)=\textbf{X}_0$ represents the initial state.
The function $\widetilde{\mathbf{F}}$ represents the vector field governing the instantaneous rate of change.
The dynamics are influenced by two distinct types of inputs:
\begin{itemize}
\item System Parameters ($\boldsymbol{\beta}$): A $p$-dimensional vector of constant parameters characterizing the individual unit (e.g., patient physiology or physical constants). We partition this vector into observed components $\boldsymbol{\beta}^{\text{obs}}$ and unobserved latent components $\boldsymbol{\beta}^{\text{U}}$, such that $\boldsymbol{\beta} = (\boldsymbol{\beta}^{\text{obs}}, \boldsymbol{\beta}^{\text{U}}) \in \mathcal{B}$, where $\mathcal{B}$ is the set of all possible constant parameters of the system.
\item Intervention ($\boldsymbol{\eta}$): An $r$-dimensional vector of time-varying external intervention $\boldsymbol{\eta}(t) \in \mathcal{H}$,  applied to the system (e.g., drug infusion rate), where $\mathcal{H}$ is the set of all possible interventions in the dynamics.
\end{itemize}

While Eq.~\eqref{eq:xue-general-form} describes the deterministic dynamics of a single unit, we require reasoning over a population. Accordingly, we treat the unit-specific tuple of the initial state, parameters, and intervention, $(\textbf{X}_0, \boldsymbol{\beta}, \boldsymbol{\eta})$, as a realization of random variables drawn from a joint distribution $\mathcal{P}$ over the configuration space $\mathcal{X} \times \mathcal{B} \times \mathcal{H}$. This probabilistic formulation establishes the basis for the independence assumptions discussed in Section~\ref{sec:neural-ode-methods}.
We demonstrate the formulation using the pendulum example provided in Box~\ref{box:pendulum-example}. 
 
\begin{figure}
    \centering
    \includegraphics[width=\linewidth]{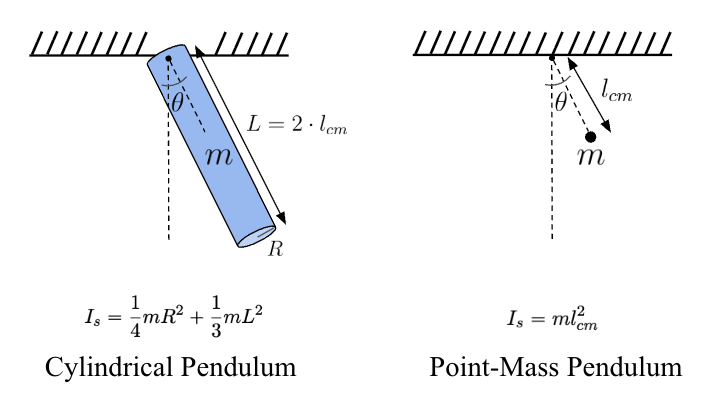}
    \caption[Pendulums sketch]{Cylindrical and point-mass pendulums}
    \label{fig:pendulums-sketch}
\end{figure}

Our goal is to estimate the state evolution $\mathbf{X}(t)$ under different interventions $\boldsymbol{\eta}(t)$, with interventions both within-distribution and out-of-distribution (OOD). We assume the true generative process is unknown, and potentially more complex than our best mechanistic approximation, and that the parameters $\boldsymbol{\beta}$ may also be unknown.

To estimate the effect of the intervention $\boldsymbol{\eta}(t)$ on the state vector $\mathbf{X}(t)$, we decompose the transition dynamics in Eq.~\eqref{eq:xue-general-form} into four parts: parametric vs. nonparametric, and intervention-dependent vs. intervention-independent components.
Specifically, we impose the following structural assumption on the transition operator:
\begin{assumption}[Additive Treatment Effect]
The transition operator governing the underlying dynamics, $\frac{d\mathbf{X}(t)}{dt}$, can be  decomposed into intervention-dependent and intervention-independent components:
\begin{align}
\widetilde{\mathbf{F}}\{t, &\textbf{X}(t), \boldsymbol{\beta}, \boldsymbol{\eta}(t)\} = \nonumber \\& \widetilde{\mathbf{F}}^\psi\{t, \textbf{X}(t), \boldsymbol{\beta}\} \nonumber  + \widetilde{\mathbf{F}}^\eta\{t, \textbf{X}(t), \boldsymbol{\beta}\} \cdot \boldsymbol{\eta}(t).
\end{align}
\end{assumption}

The functions $\widetilde{\mathbf{F}}^{\psi}$ are nuisance components, whereas $\widetilde{\mathbf{F}}^{\eta}$ represents the treatment effect. We assume that each of these functions can be further decomposed into parametric and nonparametric parts as follows: 

\begin{examplebox}{Pendulum Example}
\label{box:pendulum-example}
Consider two models of a pendulum. In the simplest case, a frictionless 2D pendulum, the dynamics are described by the following ODE:
\begin{equation*}
     I_s \frac{d^2\theta(t)}{dt^2} = -m g l_{cm} \sin\!\big(\theta(t)\big) + \tau_{\text{ext}}(t), 
\end{equation*}
where $m$ is the total mass, $g$ is the gravitational constant, $I_s$ is the moment of inertia about the fixed axis, $l_{cm}$ 
is the distance from the pivot to the center of mass, and $\tau_{\text{ext}}(t)$ denotes the total external torque acting about the fixed axis. This second-order differential equation can be rewritten as a system of first-order differential equations using the simplified version of the state vector representation described in Eq.~\eqref{eq:xue-general-form}. 
Specifically, for the case of a cylindrical rod rotating about its edge with $I_s = \tfrac{1}{4} m R^2 + \tfrac{1}{3} m L^2$ and $\boldsymbol{\beta} = \{m,L,R \}$ we obtain:  
\begin{subequations}
\label{eq:general-pendulum-first-order-cyl}
\begin{align}
    \frac{d\theta(t)}{dt} &= \omega(t) \\
    \frac{d\omega(t)}{dt} &= -\frac{ g l_{cm}}{\tfrac{1}{4}  R^2 + \tfrac{1}{3}  L^2} \sin\!\big(\theta(t)\big) + \frac{\tau_{\text{ext}}(t)}{\tfrac{1}{4} m R^2 + \tfrac{1}{3} m L^2}
\end{align}
\end{subequations}
and for the case of a point-mass pendulum with $I_s = m l_{cm}^2$ and $\boldsymbol{\beta} = \{m,l_{cm} \}$ the system becomes:
\begin{subequations}
\label{eq:general-pendulum-first-order-mass}
\begin{align}
    \frac{d\theta(t)}{dt} &= \omega(t) \\
    \frac{d\omega(t)}{dt} &= -\frac{g }{l_{cm}} \sin\!\big(\theta(t)\big) + \frac{\tau_{\text{ext}}(t)}{m l_{cm}^2} \;.
\end{align}
\end{subequations}
In both cases the state vector is $\textbf{X}(t) = \{ \theta(t) , \omega(t) \}^\top$, and the intervention $\boldsymbol{\eta}(t)$ corresponds to $\tau_{\text{ext}}(t)$. An illustrative sketch of the point-mass and cylindrical pendulums is shown in Fig.~\ref{fig:pendulums-sketch}. 
\end{examplebox}

\begin{assumption}[Hybrid Transition Operator]
Both the intervention dependent and the intervention independent components of the transition operator are composed of two additive components: a parametric mechanistic term and a nonparametric term, i.e., 
\begin{align*}
\widetilde{\mathbf{F}}^\alpha\{t, &\textbf{X}(t), \boldsymbol{\beta}\} = \\ &\mathbf{F}_p^\alpha \{t, \textbf{X}(t), \boldsymbol{\beta}\} \nonumber  + \mathbf{F}_{np}^\alpha \{t, \textbf{X}(t), \boldsymbol{\beta}\}, \quad \alpha \in \{\psi, \eta\},
\end{align*}
\end{assumption}
where $\{\textbf{F}_p^\psi, \textbf{F}^{\eta}_p\}$ and $\{\textbf{F}_{np}^\psi, \textbf{F}^{\eta}_{np}\}$ denote the parametric and nonparametric components of $\mathbf{\tilde{F}}$, respectively.

Therefore, the overall model takes the form
\begin{align}
\label{eq:s2}
    \frac{d\textbf{X}(t)}{dt} 
        &= \textbf{F}_p^\psi\{t, \textbf{X}(t), \boldsymbol{\beta}\} 
         + \textbf{F}_{np}^\psi\{t, \textbf{X}(t), \boldsymbol{\beta}\} \nonumber \\
        &\quad + \Big[\textbf{F}^\eta_p\{t, \textbf{X}(t), \boldsymbol{\beta}\} 
         + \textbf{F}^\eta_{np}\{t, \textbf{X}(t), \boldsymbol{\beta}\}\Big] \cdot \boldsymbol{\eta}(t), \nonumber \\
    \textbf{X}(t_0) &= \textbf{X}_0.
\end{align}
In addition, we add identifiability assumptions
\begin{assumption}[Mechanistic Identifiability]
The observed parameters $\boldsymbol{\beta}^{\text{obs}}$ of the mechanistic components $\mathbf{F}_{p}^{\psi}$ and $\mathbf{F}_{p}^{\eta}$ is assumed to be uniquely recoverable from data generated under the parametric model: Let $\mathbf{\mathcal{F}}_p^{\psi,\eta}: \{\boldsymbol{\beta}^\text{obs}, \textbf{X}_0, \boldsymbol{\eta}(t)\} \rightarrow \mathcal{D} \in \mathcal{X}^{[0,T_{\max}]}$ be the function that maps $\{\boldsymbol{\beta}^\text{obs}, \textbf{X}_0, \boldsymbol{\eta}(t)\}$ to data generated by the parametric model. We assume $\mathbf{\mathcal{F}}_p^{\psi,\eta}$ is injective in $\boldsymbol{\beta}^\text{obs}$. 
\end{assumption}

An additional assumption is the absence of time-independent unobserved confounding (UC)
\begin{assumption}[No Time-Independent UC]
The time dependent intervention is independent of all unobserved parameters, 
$\boldsymbol{\eta} \;\perp\!\!\!\perp\; \boldsymbol{\beta}^\text{U} \; \mid \; \textbf{X}, \boldsymbol{\beta}^\text{obs} \qquad \forall t \in [t_0, t_{\max}]$. 
\end{assumption}

Furthermore, because the intervention is time-dependent, we require no feedback condition:
\begin{assumption}[No Time-Dependent UC]
The intervention $\boldsymbol{\eta}(t') \quad \forall t' \in [t_0, t_{\max}]$ is known \emph{a priori} and is independent of states and parameters at earlier times $[t_0, t')$. $ \boldsymbol{\eta}(t') \;\perp\!\!\!\perp\; \textbf{X}(s), \boldsymbol{\beta} \qquad \forall s \in [t_0, t') , \quad  \forall t' \in [t_0, t_{\max}]$
\end{assumption}
The last two assumptions imply that we focus on interventions that, while potentially time-varying and covariate specific, are \emph{non-adaptive}: they are fixed \emph{a priori}. 

To illustrate the assumptions, consider the pendulum example. Suppose the dataset of observations is generated by a cylindrical rod rotating about its edge, while our mechanistic prior $\{\textbf{F}_p^\psi, \textbf{F}^{\eta}_p\}$ is based on the simplified point-mass pendulum model of Eq.~\eqref{eq:general-pendulum-first-order-mass}. 
Thus, 
\begin{equation*}
    \mathbf{F}_p^\psi = \begin{bmatrix} \omega \\ -\frac{g}{l_{cm}} \sin(\theta) \end{bmatrix}, 
    \quad 
    \mathbf{F}^\eta_p = \begin{bmatrix} 0 \\ \frac{1}{m l_{cm}^2} \end{bmatrix}.
\end{equation*}

Our goal is to estimate the parameter vector $\boldsymbol{\beta}=\{m, l_{cm}\}$ and the nonparametric vectors of functions $\{\textbf{F}_{np}^\psi, \textbf{F}^{\eta}_{np}\}$ in order to estimate the state vector under different types of interventions.
The nonparametric components may be implemented using neural networks of various types, where their input consists of the information known at the current time step and their output is a correction term.
We note that in the pendulum example, when $\tau_{\text{ext}} = 0$ the parameter $m$ does not influence the dynamics and is therefore not identifiable from data.

\section{Methods}
\label{sec:neural-ode-methods}
A major challenge in training a hybrid model is that, in many cases, the actual generative model and the required physical parameters, $\boldsymbol{\beta}$, or their labels in the dataset - are unknown. We observe only the time-dependent state vectors of the samples, $\mathbf{X}(t)$, in the dataset, together with the interventions, $\boldsymbol{\eta}(t)$.
Another well-known challenge in designing and training hybrid models is that enforcing parametric physical constraints often increases the error during data-driven optimization, causing the overall model to eventually ignore the constrained component and rely solely on the data-driven part \citep{takeishi_physics_2021}.  
To address these challenges, we develop the following method, illustrated in Fig.~\ref{fig:training-process} and detailed in Alg.~\ref{alg:training-proc}.

We distinguish between two settings. In some applications, the mechanistic parameters $\boldsymbol{\beta}$ are observed, as in the Propofol example (Section~\ref{sec:pk-experiment}). In other cases, such as the pendulum example, $\boldsymbol{\beta}$ is unobserved and must be inferred. For the latter, we introduce an auxiliary pretraining step based on simulated data.
Specifically, we generate a \emph{mechanistic simulated dataset} by sampling $\boldsymbol{\beta}$ from user-defined ranges and simulating trajectories under the known parametric dynamics, $\{\mathbf{F}_p^\psi, \mathbf{F}_p^\eta\}$. This dataset consists of state trajectories $\mathbf{X}(t)$, interventions $\boldsymbol{\eta}(t)$, and corresponding parameter labels $\boldsymbol{\beta}$. An encoder is then trained to map simulated trajectories to estimates of $\boldsymbol{\beta}$ using a MSE loss. The inferred parameters $\widehat{\boldsymbol{\beta}}$ should not be interpreted as estimates of the true generative parameters, but rather as effective parameters that enable the mechanistic component to provide a meaningful inductive bias when combined with the nonparametric correction.


Then, once the mechanistic component is complete - either because the true parameters are known or because they have been inferred - we freeze the mechanistic component of each sample, and the encoder weights when applicable, and train the complementary data-driven correction components $\{\mathbf{F}_{np}^\psi, \mathbf{F}_{np}^\eta\}$, conditioned on the parametric part. This allows the model to learn patterns from data without discarding the parametric contribution. Training the nonparametric part separately from the mechanistic and encoder components thereby addresses the second challenge.
Furthermore, separating the effects into intervention-related $\mathbf{F}^\eta$ and intervention-unrelated $\mathbf{F}^\psi$ components  enables more accurate estimation of the intervention effect, as opposed to relying on a fully nonparametric signal estimation where the intervention is part of the input. This, in turn, is expected to promote better intervention-related generalization. Additional design choices and considerations are provided in Appendix~\ref{sec:appx-design-choices}.

\begin{algorithm}[!ht]
\caption{Hybrid model training procedure}\label{alg:training-proc}
\begin{algorithmic}[1]
\INPUT $\{(\mathbf{X}_i(t), \boldsymbol{\eta}_i(t))\}_{i=1}^{n_{\text{train}}}$, $\{(\mathbf{X}_j(t), \boldsymbol{\eta}_j(t))\}_{j=1}^{n_{\text{test}}}$ for all $t\in[t_0,t_{\max}]$: training and test set
\INPUT $\{\boldsymbol{\eta}'_j(t)\}_{j=1}^{n_{\text{test}}}$: counterfactual interventions
\INPUT $\{\mathbf{F}_p^\psi, \mathbf{F}_p^\eta\}$: prior knowledge of $d\mathbf{X}/dt$
\INPUT (optional) $\{\boldsymbol{\beta}_i\}_{i=1}^{n_{\text{train}}}$, $\{\boldsymbol{\beta}_j\}_{j=1}^{n_{\text{test}}}$: training and test set mechanistic parameters (if known)

\IF{$\textsc{BetaUnknown}$} 
    \STATE Generate synthetic trajectories under the mechanistic model with parameter labels $\boldsymbol{\beta}$.
    \STATE Train an encoder $En$ to predict $\boldsymbol{\beta}$ from trajectories, e.g. by minimizing the $MSE(\boldsymbol{\beta}, \widehat{\boldsymbol{\beta}})$.
    \STATE For each training sample $i$, set $\widehat{\boldsymbol{\beta}}^{\,i} \leftarrow En(\mathbf{X}_i(\cdot), \boldsymbol{\eta}_i(\cdot))$.
\ELSE 
    \STATE For each training sample $i$, set $\widehat{\boldsymbol{\beta}}^{\,i} \leftarrow \boldsymbol{\beta}_i$.
\ENDIF

\STATE Train the hybrid Neural ODE by fitting $\mathbf{F}_{np}^\psi$ and $\mathbf{F}_{np}^{\eta}$
       conditioned on $\widehat{\boldsymbol{\beta}}^{\,i}$, $\boldsymbol{\eta}_i(t)$, $\mathbf{F}_p^\psi$, and $\mathbf{F}_p^\eta$,
       e.g., by minimizing reconstruction MSE.
\IF{$\textsc{BetaUnknown}$} 
    \STATE For each test sample $j$, use the beginning of the sequence to predict $\widehat{\boldsymbol{\beta}}^{\,j} \leftarrow En(\mathbf{X}_j(\cdot), \boldsymbol{\eta}_j(\cdot))$
\ELSE 
    \STATE For each test sample $j$, set $\widehat{\boldsymbol{\beta}}^{\,j} \leftarrow \boldsymbol{\beta}_j$.
\ENDIF
\STATE For each test sample $j$, use the hybrid model to predict the counterfactuals $\mathbf{X}'_j(t)$ under $\boldsymbol{\eta}'_j(t)$ given $\widehat{\boldsymbol{\beta}}^{\,j}$.
\OUTPUT $\{\mathbf{X}'_j(t)\}_{j=1}^{n_{\text{test}}}$: outcomes under counterfactual interventions
\end{algorithmic}
\end{algorithm}

\section{Experiments}
\label{sec:neural-ode-exp}

We demonstrate our approach through two case studies.
For each, we employ a complex generative process as the ground truth, while a simpler, incomplete model represents the available prior knowledge used as the integrated parametric component. The first case is a pendulum system with periodic dynamics and a steady intervention, where the observed data are time-dependent state trajectories, and the physical parameters are unknown and must be inferred. The second is a model of the PK of Propofol, in which patient parameters are given, and the objective is to select the total dose, administered as a sequence of bolus infusions, to reach a target effect-site level. These cases cover different types of dynamics, with different data-availability scenarios and analysis goals, all of which are common in practice. 

We analyze both case studies through the lens of three approaches: purely mechanistic models expressed through physical ODEs, purely data-driven models implemented with Neural ODE \citep{chen_neural_2018}, and our hybrid framework, which uses physical ODEs as a mechanistic prior integrated with correction networks using a Neural ODE architecture. Our nonparametric networks are implemented using Multilayer Perceptrons (MLPs). Additional details are provided in Appendices \ref{sec:appx-pendulum-details}-\ref{sec:appx-pk-details}.  

\subsection{Pendulum Experiments}
\label{sec:pendulum-experiments}
Consider a case in which the data describe the current state of a rigid cylinder rotating about its edge. Our prior knowledge of pendulum dynamics is represented by the simplified point-mass pendulum. Both scenarios are illustrated in Fig.~\ref{fig:pendulums-sketch}. 
The goal of this experiment is to evaluate how effectively the proposed hybrid models leverage imperfect mechanistic knowledge, together with data generated from the true generative model, to predict system trajectories under observed interventions. Here, test samples differ from training samples not by reconstructing previously observed trajectories, but by predicting trajectories for new initial conditions, parameter values, and interventions that were not encountered during training.

We generate a dataset consisting of $n=5000$ observations, each with state vectors $\{\theta(t), \omega(t)\}$ that follow the dynamics of cylinder rotating about its edge described in Eq.~\eqref{eq:general-pendulum-first-order-cyl}.
We begin by sampling the true parameters $\{m, L, R\}$ of the cylinder such that $m \sim \mathcal{U}(3.5,4)$ kg, $L \sim \mathcal{U}(4,4.5)$ meter, $R \sim \mathcal{U}(2,2.5)$ meter, 
and the initial conditions $\{\theta_0, \omega_0\}$ with $\theta_0 \sim \mathcal{U}(-0.2, 0.2)$ rad and $\omega_0 \sim \mathcal{U}(-0.1, 0.1)$ rad/sec.

For each observation, we apply a constant intervention such that, for all $t$, the intervention magnitude satisfies $\tau_{\text{ext}}(t) = 10 + 0.5\,k$, where $k \sim \mathcal{U}\{0,\ldots,4\}$. The state trajectories are then constructed over the time interval $t \in \{0, \ldots, t_{\max}\}$, with $t_{\max} = 10$, using a Runge-Kutta integration method and a step size of $\Delta t = 0.1$. 
We generate two types of test sets in a similar fashion, with the two differing only in the intervention distribution: one is \textbf{in-distribution} (same as training set) and one is \textbf{out-of-distribution} with $k \sim \mathcal{U}\{5,6,7\}$.

The known prior knowledge is given as the point-mass pendulum equation system described in Eq.~\eqref{eq:general-pendulum-first-order-mass}, with unknown parameters that must be inferred.
Thus, we include an encoder  based on Eq.~\eqref{eq:general-pendulum-first-order-mass} to infer the parameters of the mechanistic model $\{m, l_{cm}\}$.

The hybrid model follows the point-mass prior with correction networks, i.e.,
\begin{align}
\label{eq:hybrid-pendulum}
    \frac{d\theta(t)}{dt} =& \omega(t) + \textcolor{Blue}{F_{np}(t, \theta(t), \omega(t), m, l_{cm})} \nonumber \\
        &+ \textcolor{Blue}{F_{np}^{\eta}(t, \theta(t), \omega(t), m, l_{cm})} \cdot \tau_{ext}(t), \nonumber \\
    \frac{d\omega(t)}{dt} =& -\frac{g}{l_{cm}} \sin\!\big(\theta(t)\big) + \frac{\tau_{ext}(t)}{m l_{cm}^2 } \nonumber \\
        &+ \textcolor{OliveGreen}{G_{np}(t, \theta(t), \omega(t), m, l_{cm})} \nonumber \\ 
        &+ \textcolor{OliveGreen}{G_{np}^{\eta}(t, \theta(t), \omega(t), m, l_{cm})} \cdot \tau_{ext}(t).
\end{align}
where $\{\textcolor{Blue}{F_{np}}, \textcolor{Blue}{F_{np}^\eta}, \textcolor{OliveGreen}{G_{np}}, \textcolor{OliveGreen}{G_{np}^\eta}\}$ are correction networks learned from data. 
Lastly, we compared the results of the mechanistic only and hybrid models to a fully data-driven baseline model as defined in Eq.~\eqref{eq:xue-general-form}.

In our implementation, the encoder and the networks denoted by $F_{np}, F_{np}^\eta, G_{np}, G_{np}^\eta$ and the fully data-driven networks are all multilayer perceptrons (MLPs). These networks parameterize the right-hand side of the system’s differential equation. Given initial conditions, $\frac{d\textbf{X}(t)}{dt}$ is integrated using a numerical ODE solver to generate the state trajectories up to $T_{\max}$. During training, we minimize a reconstruction MSE, defined as the squared difference between the reconstructed and observed trajectories, by optimizing the network parameters \citep{chen_neural_2018}.

\subsubsection{Test Reconstruction Performance}
\begin{figure*}
    \centering
    \includegraphics[width=0.8\linewidth]{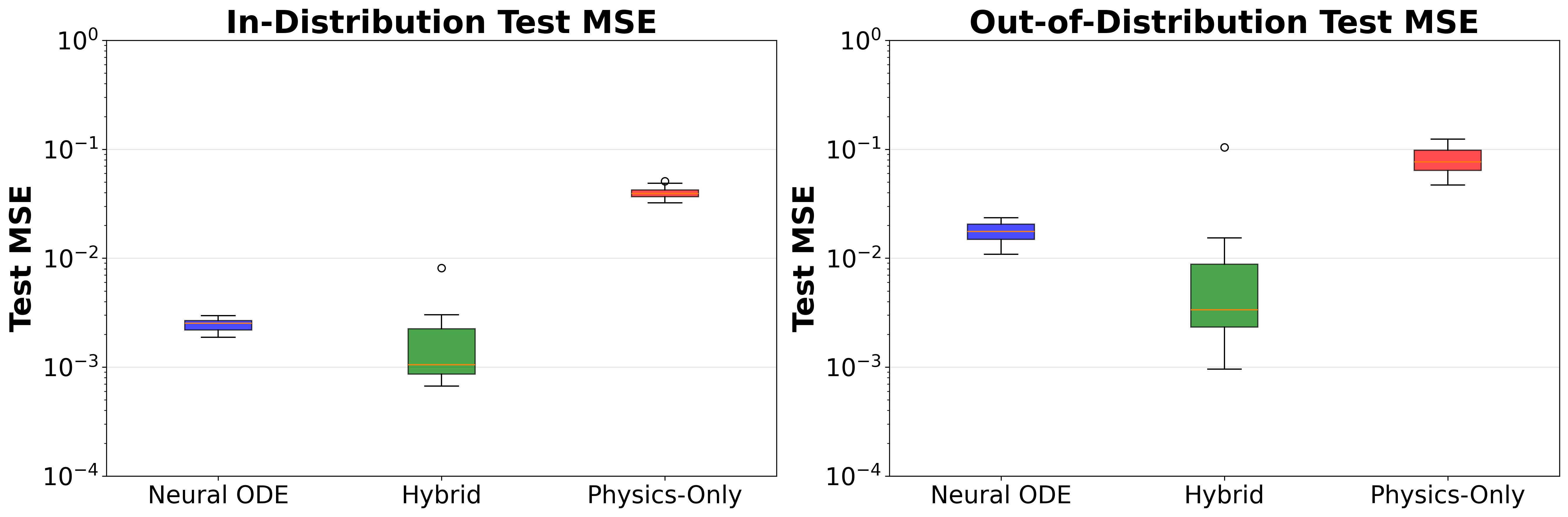}
    \caption[Pendulum mean reconstruction MSE]{\textbf{Test Reconstruction Results on Pendulum}: Box-plot across 20 replications of the mean reconstruction MSE (averaged over samples and time) for 50{,}000 test trajectories under \textbf{in-distribution} and \textbf{out-of-distribution} intervention types, on a logarithmic scale. Here, reconstruction refers to forecasting the system trajectory from randomly sampled initial conditions and parameters drawn from the training distribution, under given interventions (in-distribution or out-of-distribution), using the learned models, and comparing the resulting trajectories to those generated by the true underlying dynamics.}
    \label{fig:test_mse_replications_summary}
\end{figure*}

A box-plot of the mean test reconstruction MSE, for the three models, based on 20 replications of the experiment, are shown in log-scale in Fig.~\ref{fig:test_mse_replications_summary}. 
Evidently, relying solely on the simplified mechanistic prior knowledge - namely, the physics-only model - yields the worst performance, as expected, due to its incomplete physics. The fully data-driven and hybrid models are comparable in-distribution, with a slight advantage for the hybrid model. In the OOD regime, all models exhibit higher MSE comparing to their results in-distribution, as expected; nevertheless, the hybrid model maintains its performance better than the fully data-driven model and achieves the best results.

\subsubsection{Intervention Outcome Prediction}
\begin{figure}
    \centering
    \includegraphics[width=\linewidth]{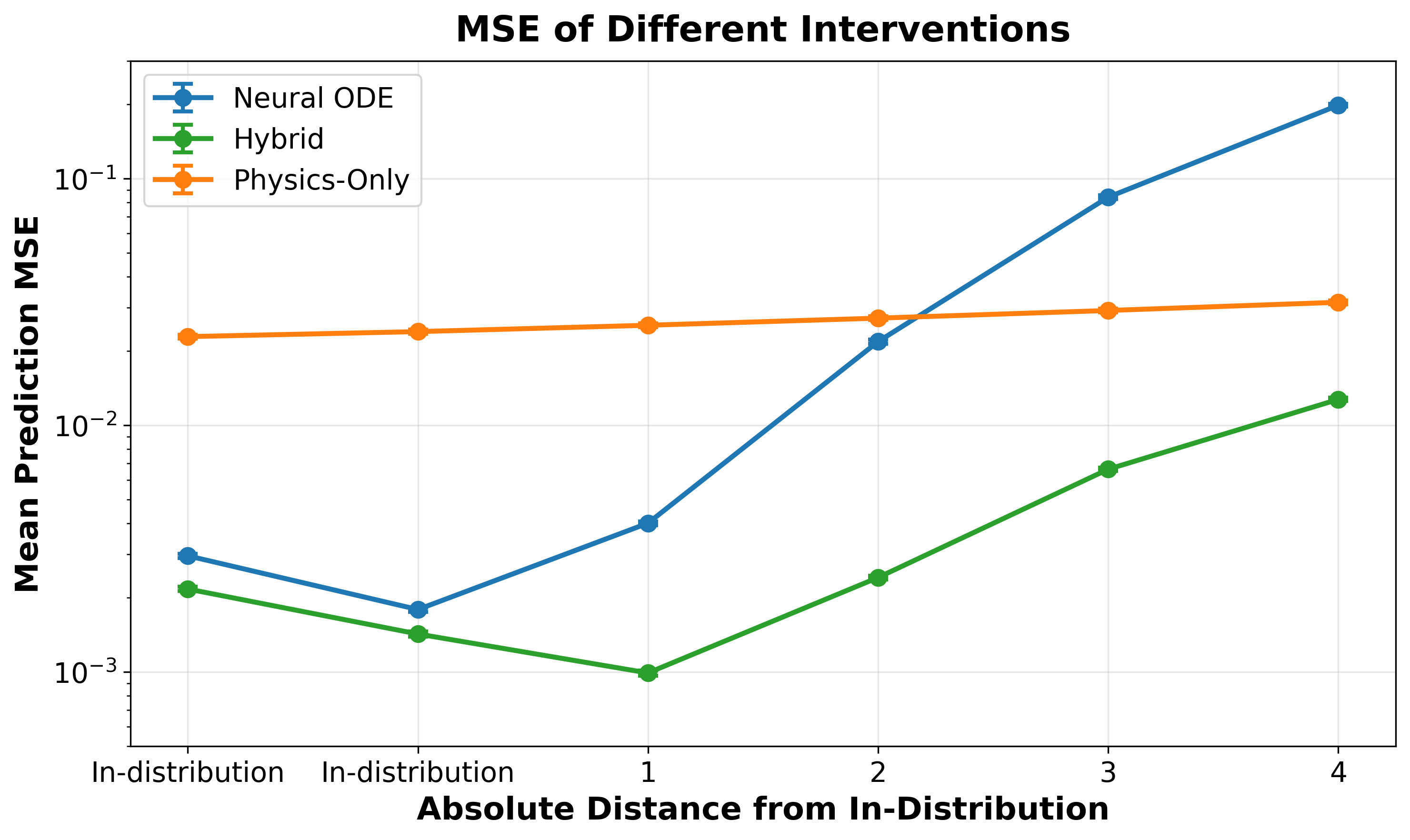}
    \caption[Different Intervention Outcomes MSE]{\textbf{Intervention Outcome Prediction Pendulum}: Mean MSE and SEM across 5,000 test samples for different intervention values for all three models as a function of distance from training intervention set (logarithmic scale).}
    \label{fig:interventions-outcomes}
\end{figure}

Next, we evaluate the models' ability to predict intervention outcomes to facilitate optimal intervention selection. We generate 5,000 observations that follow the cylinder dynamics with $T_{\max} = 20$ seconds and $\Delta t=0.1$ second. In the first 100 time points (up to $t = 10$), we apply an intervention of $\tau_{\text{ext}}(t) = 10$. At $t = 10$, we generate a set of counterfactuals for each observation using $\tau_{\text{ext}}(t) = 6 + k$, $k=0,\ldots,5$. We then compare the models' ability to predict these outcomes, where the interventions $\tau_{\text{ext}}(t) \in \{10,11\}$ are considered in-distribution and $\tau_{\text{ext}}(t) \in \{6,\ldots,9\}$, are considered OOD. In this experiment, the first 100 time points are used by the encoder to infer the unknown mechanistic parameters. The subsequent 100 time points are withheld from the models and serve as the ground truth for evaluating intervention-outcome predictions.
The mean intervention outcome prediction MSE for the three models, across all counterfactual values, is shown in Fig.~\ref{fig:interventions-outcomes}. The fully data-driven and hybrid models are comparable in-distribution. However, similarly to the reconstruction results, when moving further beyond previously seen interventions, i.e., deeper into the OOD regime, the hybrid model maintains its performance better than the fully data-driven model.


\subsection{Pharmacokinetics Experiment}
\label{sec:pk-experiment}

PK models describe the time evolution of drug concentration in the body using differential equations that account for drug absorption, transport, and clearance. A standard mathematical representation is the three-compartment mammillary model presented in Fig.~\ref{fig:3comp-model}. 
The central compartment ($V_1$) represents the blood and highly perfused tissues, while two peripheral compartments ($V_2, V_3$) represent muscle and adipose tissue, respectively. The dynamics are described by:
\begin{equation}
\label{eq:pk_model}
\begin{aligned}
\frac{dA_1}{dt} &= -(k_{10} + k_{12} + k_{13})A_1 + k_{21}A_2 + k_{31}A_3 + u(t) \\
\frac{dA_c}{dt} &= k_{1c}A_1 - k_{c1}A_c \quad , \quad c=2,3
\end{aligned}
\end{equation}
where $A_i$ denotes the amount of drug in compartment $i$, $k_{ij}$ represents the transfer rate from compartment $i$ to $j$, and $u(t)$ is the infusion rate.

To link concentration to clinical effect, e.g., loss of consciousness, an effect-site compartment is added. The effect-site concentration ($C_e$) lags behind the plasma concentration ($C_p = A_1/V_1$) according to a first-order delay governed by the rate constant $k_{e0}$:
\begin{equation}
\label{eq:effect_site_deriv}
\frac{dC_e}{dt} = k_{e0}(C_p - C_e).
\end{equation}
The clinical goal is to target a specific effect level, e.g. reduction of brain activity by 50\%, by achieving the required concentration at the effect site $C_e$.

Propofol is the most widely used intravenous anesthetic agent for the induction and maintenance of general anesthesia. The model of \cite{schnider_influence_1998} parameterizes a three-compartment PK structure using patient covariates such as age, height, weight, and lean body mass. 
Owing to the practical difficulty of conducting interventional studies in humans, the model was developed from a small cohort of $n=24$ healthy adults. 
The linear covariate structure fails to capture nonlinear physiological scaling effects, leading to biased predictions, particularly in high body mass index (BMI) and extreme-age regimes. 
The \cite{eleveld_pharmacokinetic_2018} model is a unified population PK model for Propofol, developed on a much larger cohort ($n=1000$) spanning wide age and BMI ranges. It uses allometric scaling, maturation functions, and rich covariate effects, yielding improved accuracy across pediatric and adult populations.
Accordingly, in this experiment, \cite{eleveld_pharmacokinetic_2018} is used as the generative model, while the \cite{schnider_influence_1998} model serves as simplified prior. 

In modern practice, PK models such as \cite{schnider_influence_1998} are embedded in Target Controlled Infusion (TCI) pumps. The system solves an inverse problem: given a clinician-set target $C_e$ and patient characteristics, the pump computes the required infusion dose, delivered as bolus injections every several seconds until the total dose is administered. The accuracy of this intervention, and its ability to achieve the anesthetic state without missing the target or overshooting and causing adverse effects, depend critically on the fidelity of the underlying model \citep{enlund_tci_2008}. Our goal is to compare the mechanistic, hybrid, and fully data-driven models in their ability to function as such controllers to give the correct dosing schedule.


So far, we used observations with randomly sampled, uncorrelated covariates. In practice, however - especially in healthcare - patient covariates are often correlated through both known and unknown relationships. Therefore, we consider an experimental setting with realistic covariates, along with simulated treatments and outcomes.
Specifically, we consider $N=12,155$ patients from the MIMIC-IV database \citep{johnson_mimic-iv_2022, goldberger_physiobank_2000} who have records of Propofol administration during their admission. For each patient we extract age, sex, weight, height, and a binary indicator of opioid treatment prior to the first record of Propofol. 
The target concentration at the effect-site, denoted $C_e^*$, is known to be age dependent \citep{schnider_influence_1999} and here is simulated as $C_e^*(\text{age}) = 4 - 0.04(\text{age} - 18)$. Initial Propofol concentrations are zero in all compartments. Infusion is delivered as a series of $30\,\mathrm{mg}$ boluses every $10$ seconds until a total dose $D$ is reached. The optimal dose $D^*$ is chosen by simulating age-dependent candidate doses and selecting the one whose trajectory yields an effect-site concentration closest to $C_e^*$. The candidate dose set is weight dependent and is defined as $D(\text{weight}) = \text{weight} \cdot d$, where the normalized dose values $d$ are similar to the FDA Propofol guidelines \citep{fda_propofol_2017}. Specifically,
$d \in \{1.5,\ldots,2.5\}\,\frac{\mathrm{mg}}{\mathrm{kg}}$ with step $0.1\,\frac{\mathrm{mg}}{\mathrm{kg}}$ for patients younger than 55 and $d \in \{1,\ldots,1.5\}\,\frac{\mathrm{mg}}{\mathrm{kg}}$ with step  $0.1\,\frac{\mathrm{mg}}{\mathrm{kg}}$ for patients aged 55 or older. Ground-truth synthetic trajectories are generated using the oracle model of \cite{eleveld_pharmacokinetic_2018} with $T_{\max}=210 \, \mathrm{seconds}$ and $\Delta t = 0.5 \, \mathrm{seconds}$. An example of the intervention and resulting trajectories is shown in Fig.~\ref{fig:patient_counterfactuals_example}. 
We randomly hold out 25\% of the data as a test set. In addition, patients with BMI $>30$ or age $>60$ are excluded from the training set and instead form an OOD population used exclusively for testing. Overall, we consider three test groups: in-distribution (held out, but from the same training set distribution); OOD (age $>60$ or BMI $>30$); and extreme OOD (age $>60$ and BMI $>30$). Note that OOD is defined with respect to the training distribution of the neural networks only. The mechanistic model of \cite{schnider_influence_1998} was estimated using $n=24$ patients aged 26--81.

As the mechanistic model we use the three compartment model, i.e. Eq.~\eqref{eq:pk_model}-\eqref{eq:effect_site_deriv}, with the parameters of the \cite{schnider_influence_1998} model. In our notation defined in Eq.~\eqref{eq:s2} these are $\textbf{F}_p^\psi$ and $\textbf{F}_p^\eta$, while the intervention $u(t)$ is $\boldsymbol{\eta}(t)$. Due to the model structure, where $A_2$ and $A_3$ interact linearly only with $A_1$ and not with each other, and because the intervention is applied directly at $A_1$ rather than on the other compartments, we include correction networks only in the $A_1$ dynamics. This is expected to capture the full correction needed for the plasma concentration $C_p$. Additionally, we add correction networks to the effect-site concentration $C_e$. Given its original mechanistic structure, we allow corrections to both the time constant and an additional additive term. Thus, the hybrid components are:
\begin{equation}
\label{eq:pk_hybrid_model}
\begin{aligned}
\frac{dA_1}{dt} &=  \left( \frac{dA_1}{dt} \right)_{p} + F_{np}^\psi(t, \textbf{A}, Ce, \boldsymbol{\beta}) \\ 
&\qquad \qquad \qquad + F_{np}^\eta(t, \textbf{A}, Ce, \boldsymbol{\beta}) \cdot u(t)  \\
\frac{dC_e}{dt} &= \left( \frac{dC_e}{dt}\right)_{p} \cdot G_{np}(t, \textbf{A}, Ce, \boldsymbol{\beta}) + G_{np}^\psi(t, \textbf{A}, Ce, \boldsymbol{\beta})
\end{aligned}
\end{equation}
together with the mechanistic $\frac{dA_2}{dt}$ and $\frac{dA_3}{dt}$.  We ensure the output of $G_{np}$ is positive, and $(\cdot)_p$ denotes the parameteric components.
Lastly, the fully data-driven model is as defined in Eq.~\eqref{eq:xue-general-form}.
In our example, $\{F_{np}^\psi, F_{np}^\eta, G_{np}, G_{np}^\psi\}$ and the fully data-driven model networks are all MLPs.

We train the correction networks by minimizing the MSE reconstruction loss of the predicted plasma ($C_p$) and effect-site ($C_e$) concentrations. Training is done using batches of time windows of $40 \, \mathrm{seconds}$. Here, the intervention is sparse and the initial conditions are zero (and do not return to zero within the experiment time frame). Therefore, we ensure that at least $15\%$ of each batch starts at $t=0$, and additional $15\%$ of each batch includes non-zero intervention. The remaining starting points of the windows in the batch are sampled uniformly at random.

\begin{figure}
    \centering
    \includegraphics[width=0.8\linewidth]{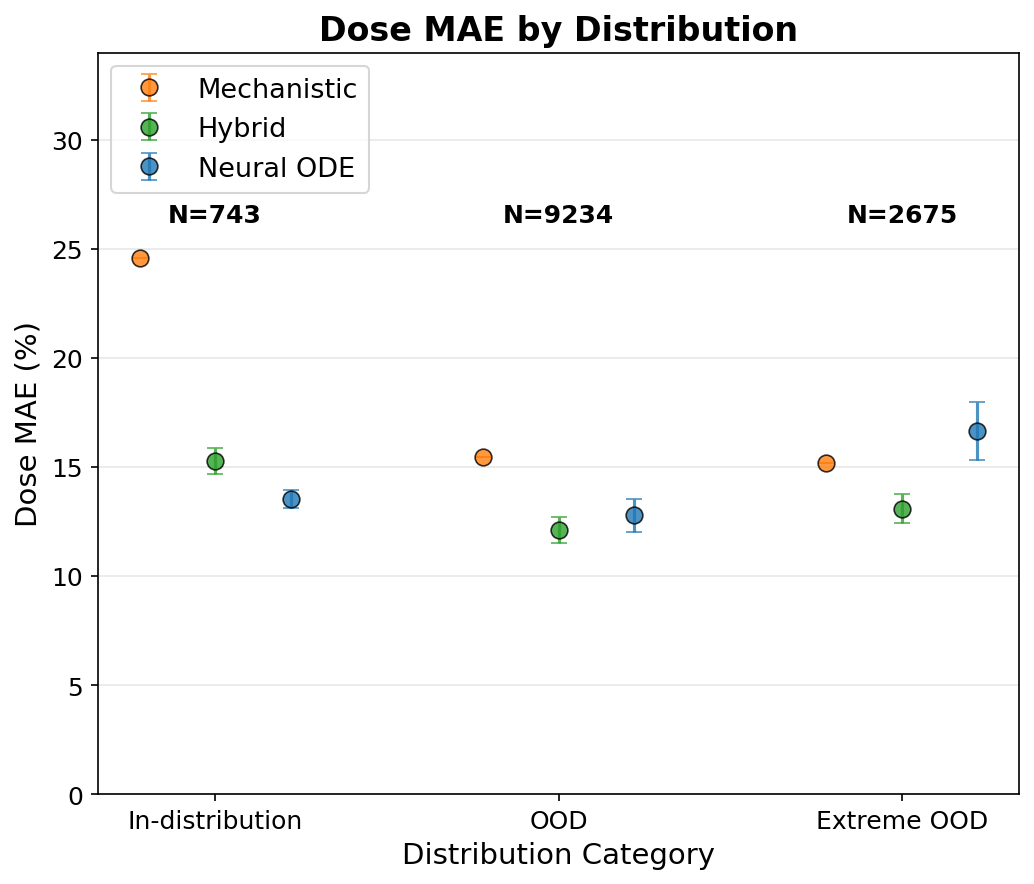}
    \caption[Selected Dose MAPE]{\textbf{PK Test Results}: Selected-dose MAPE: mean and SEM over 20 replications for three test groups: In-distribution, OOD (age $>60$ \textbf{or} BMI $>30$), and extreme OOD (age $>60$ \textbf{and} BMI $>30$). OOD is defined with respect to the networks' training distribution only. The mechanistic model of \cite{schnider_influence_1998} was estimated using 24 patients aged 26-81.} 
    \label{fig:replications_dose_ape}
\end{figure}

\paragraph{Results}
We assess the clinical utility of the models by evaluating their ability to infer the optimal induction bolus dose. For each test patient, we use each model to generate personlized counterfactual trajectories.
We select the normalized dose $d$ that yields an effect-site concentration $C_e$ closest to the target $C_e^*$. For each patient, we then compute the absolute percentage error (APE) between the selected dose $d$ and the optimal dose $d^*$ that achieves $C_e^*$ according to the oracle model of \cite{eleveld_pharmacokinetic_2018}. The mean APE (MAPE) is computed across patients within each distribution group (in-distribution, OOD, and extreme OOD). The training and evaluation procedure is repeated 20 times, and the mean and SEM of the selected dose MAPE across replications are reported in Fig.~\ref{fig:replications_dose_ape}. The fully data-driven model achieves the best performance in-distribution. However, as we move to OOD and extreme OOD settings, the hybrid model exhibits greater robustness, with consistently lower error. The mechanistic model yields the highest error overall, except in the extreme OOD regime, where the fully data-driven model performs worst.

\section{Discussion}
\label{sec:neural-ode-discussion}

We discuss hybrid mechanistic-data-driven framework for estimating outcomes under predefined, time-dependent interventions in dynamical systems with incomplete mechanistic knowledge. The core modeling idea is representing the transition operator as a mechanistic component plus learned corrections, and we explicitly separate intervention-independent from intervention-dependent effects. This decomposition is intended to preserve mechanistic anchors while letting data-driven modules absorb misspecification, focusing on counterfactual prediction.

The two case studies illustrate the same general template instantiated under different, practically motivated constraints. In the pendulum setting, mechanistic parameters are unobserved. We proposed a two-stage procedure: an encoder is pre-trained on synthetic trajectories generated from the mechanistic prior and then used to infer parameters for the real dataset before learning the corrections. In contrast, in the PK setting the relevant patient characteristics are available, so no parameter inference is needed, and modeling decisions are driven by the mechanistic structure and by the intervention (bolus dosing). Concretely, corrections are applied where they can most directly compensate for known deficiencies (e.g., plasma/effect-site components).
In both cases, we demonstrate that the hybrid approach outperforms both the purely data-driven and purely physics-based models, and hybrid modeling is most helpful when extrapolating beyond the training distribution of interventions or covariates - which is inherent in causal inference. 





\bibliographystyle{plainnat}
\bibliography{refs}

\clearpage

\appendix

\section{Pendulum Experiments - Additional Details}
\label{sec:appx-pendulum-details}

This appendix provides additional details on the model architectures and training procedures for all models used in the experiments.


\subsection{Physics Only / Encoder Architecture and Training}
\label{sec:encoder}

The encoder estimates the physical parameters of the mass pendulum dynamics $\beta=(m,l_{cm})$ from a trajectory with a given intervention $\{\theta_i(t), \omega_i(t), \tau_{\mathrm{ext}}(t)\}$.

The network architecture is a fully connected stack with:
\begin{itemize}
    \item Batch normalization of the flattened input,
    \item Linear layer $(T_{\mathrm{enc}} d) \rightarrow H$, batch normalization,
          ReLU, 
    \item Two additional hidden linear layers of size $H \rightarrow H$ with batch normalization, ReLU, 
    \item Linear layer $H \rightarrow H/2$ with batch normalization, ReLU, 
    \item Linear output layer $(H/2) \rightarrow 2$ producing $(\widehat{m},\widehat{l}_{cm})$.
\end{itemize}
The hidden dimension was set to $H=128$.

For training the encoder we used the synthetic simulated data generated as described in Section~\ref{sec:pendulum-experiments}, together with labels corresponding to the true physical parameters $\beta = (m, L)$ for each sample. The dataset is randomly split into $80\%$ training and $20\%$ validation sets. The encoder is trained using the Adam optimizer with learning rate $10^{-3}$, batch size $B=256$, weight decay $10^{-5}$, and mean-squared error loss
\[
\mathcal{L}_{\mathrm{enc}}
= \frac{1}{B}\sum_{i=1}^B \|\widehat{\beta}_i - \beta_i\|_2^2 .
\]
We employ a learning-rate reduction on plateau with mode='min', factor=0.5, patience=5, and early stopping with tolerance $\mathrm{tol}=10^{-4}$ over $25$ epochs.

\subsection{Multilayer Perceptron}

Multilayer perceptron (MLP) networks are used for both the hybrid correction networks and the data-driven model.
Each residual block consists of a sequence of LayerNorm, linear layers (from hidden dimension to hidden dimension), and Tanh activations. The output of each block is added to the input via a skip connection.
The MLP architecture includes a linear projection from the input dimension to the hidden dimension, followed by a sequence of $r$ residual blocks, a final normalization layer, and an output projection layer.

\subsection{Data-Driven Neural ODE}
\label{sec:details-data-driven-neural-ode}

The data-driven Neural ODE is composed of two MLPs: one for $\frac{d\theta}{dt}$ and one for $\frac{d\omega}{dt}$, with a hidden dimension of $d = 128$, $r = 6$ residual blocks, and Xavier initialization with $\mathrm{gain} = 1$.

For trajectory reconstruction, we use \texttt{odeint} with the \texttt{RK4} method. Training minimizes the trajectory prediction error:
\begin{equation}
\label{eq:appx-mse-loss}
\mathcal{L}_{\mathrm{ode}}
= \frac{1}{BT} \sum_{i,t}
\bigl\|
(\widehat{\theta}_{i,t}, \widehat{\omega}_{i,t})
-
(\theta_{i,t}, \omega_{i,t})
\bigr\|_2^2,
\end{equation}
using the following Neural-ODE training configuration: batch size 64, learning rate $10^{-3}$, and early stopping after 12 epochs with no improvement in validation performance.

\subsection{Hybrid Mechanistic-Data-Driven Model}

The hybrid model augments the mechanistic pendulum equations with learned residual corrections and uses the encoder to estimate $(m,l_{cm})$.
The nonparametric correction components are MLP networks with hidden dimension of $d=64$, $r=4$ residual blocks, and Xavier initialization with \( \mathrm{gain} = 0.3 \).

The model is trained to minimize the trajectory reconstruction loss described in Eq.~\eqref{eq:appx-mse-loss} using the Adam optimizer with a learning rate of $2 \times 10^{-4}$, a batch size of $64$, and early stopping after 12 epochs with no improvement in validation performance.

\clearpage

\section{Pharmacokinetics Experiment}
\label{sec:appx-pk-details}

\subsection{Models}

\subsubsection{Mechanistic Baseline \cite{schnider_influence_1998}}
In this experiment, patient's parameters are given in the data. Thus no encoder is required.
The mechanistic baseline uses the \cite{schnider_influence_1998} model parameters. This model is known to have limitations in obese populations due to its reliance on the James equation for Lean Body Mass (LBM), which can produce inaccurate results at high BMIs.

\subsubsection{Hybrid Model (Schnider + Learned Residual)}
The hybrid model augments the \cite{schnider_influence_1998} mechanistic structure with MLP neural correction networks with hidden dimension of $d=64$, $r=4$ residual blocks, and Xavier initialization with \( \mathrm{gain} = 0.2 \).

\subsubsection{Data-driven Neural ODE}
The fully data-driven baseline utilizes a 2-state formulation ($C_p, C_e$), learning the dynamics via two separate MLP networks with  $d=64$, $r=4$ residual blocks, and Xavier initialization with \( \mathrm{gain} = 1 \).

\subsection{Training and Evaluation Protocol}

To balance the contributions of plasma ($C_p$) and effect-site ($C_e$) concentrations, we use a relative Mean Squared Error (MSE) normalized by clinically relevant thresholds:

\begin{equation}
    \mathcal{L} = \frac{1}{N} \sum \frac{(C_{p, \text{pred}} - C_{p, \text{true}})^2}{25.0^2} + \frac{1}{N} \sum \frac{(C_{e, \text{pred}} - C_{e, \text{true}})^2}{3.5^2}
\end{equation}

where $25.0$ $\mu$g/mL is the safety limit for $C_p$ and $3.5$ $\mu$g/mL is the typical target for $C_e$.
Training is performed using random 40-second windows ($80$ timesteps).
We set number of training epochs to 100, batch size to 32, and learning rate to $10^{-3}$

Models are evaluated on their ability to select an optimal induction bolus. The selection criteria are prioritized as follows:
\begin{enumerate}[topsep=0pt,itemsep=0pt]
    \item \textbf{Safety:} $C_p < 25.0$ $\mu$g/mL.
    \item \textbf{Target Achievement:} Reach $C_{e, \text{goal}}(\text{age})$.
    \item \textbf{Precision:} Minimize $| \max(C_e(t)) - C_{e}^* |$.
\end{enumerate}

Performance is quantified using Median Absolute Performance Error (MDAPE):
\begin{equation}
    \text{APE} = \frac{|d - d^*|}{d^*} \times 100
\end{equation}

\clearpage

\section{Practical Considerations and Design Choices}
\label{sec:appx-design-choices}

In this appendix, we describe the technical and methodological lessons learned during the construction and training of hybrid mechanistic-data-driven networks for intervention outcome prediction. 

\subsection{Architectural Considerations}

In developing hybrid models for time-series intervention outcome prediction, beyond the architecture presented in the paper, we identified additional principles to consider:
\begin{itemize}
    \item \textbf{When are hybrid models most beneficial:} Hybrid models operate within a specific ``grey area" of knowledge. If mechanistic knowledge is too poor, requiring massive corrections, a \textbf{fully data-driven} approach may perform better by avoiding the bias of incorrect priors. Conversely, if the physics is already well-described, adding data-driven components may introduce noise and harm prediction. The hybrid approach is most beneficial when there is relevant but incomplete working knowledge that can be refined to present a full picture.
    
    
    \item \textbf{Efficiency of Simple Architectures:} Despite the complexity of the underlying physics, Multi-Layer Perceptrons (MLPs) proved to be simple yet powerful tools for predicting required corrections. When conditioned on the current state and the known mechanistic parameters, MLPs effectively captured degrees of freedom which were missing in the simplified physical model. 
\end{itemize}

\subsection{Training Procedure}

During the training procedure, additional considerations should be taken into account:

\subsubsection{Window-Based Training}

Training the model on sub-sequences (windows) instead of full trajectories proved essential for the convergence. Shorter windows minimize the requirement for long-range extrapolation during the early stages of training, reduce the computational overhead associated with numerical integration, and reduce the chance of convergence to an average constant. However, the following considerations should be taken into account:

\begin{itemize}
    \item \textbf{Intervention Sparsity:} Because interventions are often sparse (e.g., discrete bolus doses), a multiplicative correction is multiplied by zero most of the time. This leads to ``dead'' gradients, making it challenging for the intervention-dependent network to learn. We address this issue through batch composition, ensuring that each batch contains a certain proportion of nonzero interventions. 
    \item \textbf{Zero Initial Conditions:} Initial conditions are sometimes zero before interventions and do not return to zero afterward. When using window-based training, if the starting point is sampled uniformly over the time grid, zero initial conditions are almost never selected. As a result, zero initial conditions become effectively OOD, leading to large inference-time errors, since such conditions are rarely observed during training. We address this issue by enriching each batch with at least a certain proportion of samples that start from zero initial conditions.
    \item \textbf{Time constant:} In our pendulum example, where the dynamics are cyclic, small windows were sufficient to capture the governing laws. In our PK example, however, longer windows were required to capture the physical changes and train thecorrection networks. The known physics can be used for selecting an appropriate window size.
\end{itemize}



\subsubsection{Initialization and Balancing}
Lastly, we highlight additional design choices that facilitated the convergence of the correction networks:

\begin{itemize}

    \item \textbf{Weight Initialization:} We found it helpful to start with a small non-zero initialization for the correction networks (using a non-default gain hyper-parameter). This allows the mechanistic model to lead the initial training phase while the ML component gradually introduces corrections.
    
    \item \textbf{Balance Hyper-parameters:} A dedicated weight hyper-parameter was used to balance the initial contribution of the physics-based components and the corrections, which was helpful for achieving convergence.
\end{itemize}

\clearpage

\renewcommand{\thefigure}{S\arabic{figure}}
\renewcommand{\thetable}{S\arabic{table}}

\makeatletter
\@ifundefined{theHfigure}{}{%
  \renewcommand{\theHfigure}{S.\arabic{figure}}%
}
\@ifundefined{theHtable}{}{%
  \renewcommand{\theHtable}{S.\arabic{table}}%
}
\makeatother

\setcounter{figure}{0}
\setcounter{table}{0}

\begin{figure*}
    \centering
    \includegraphics[
        width=\linewidth,
        trim=3cm 1.5cm 3cm 1cm, 
        clip
    ]{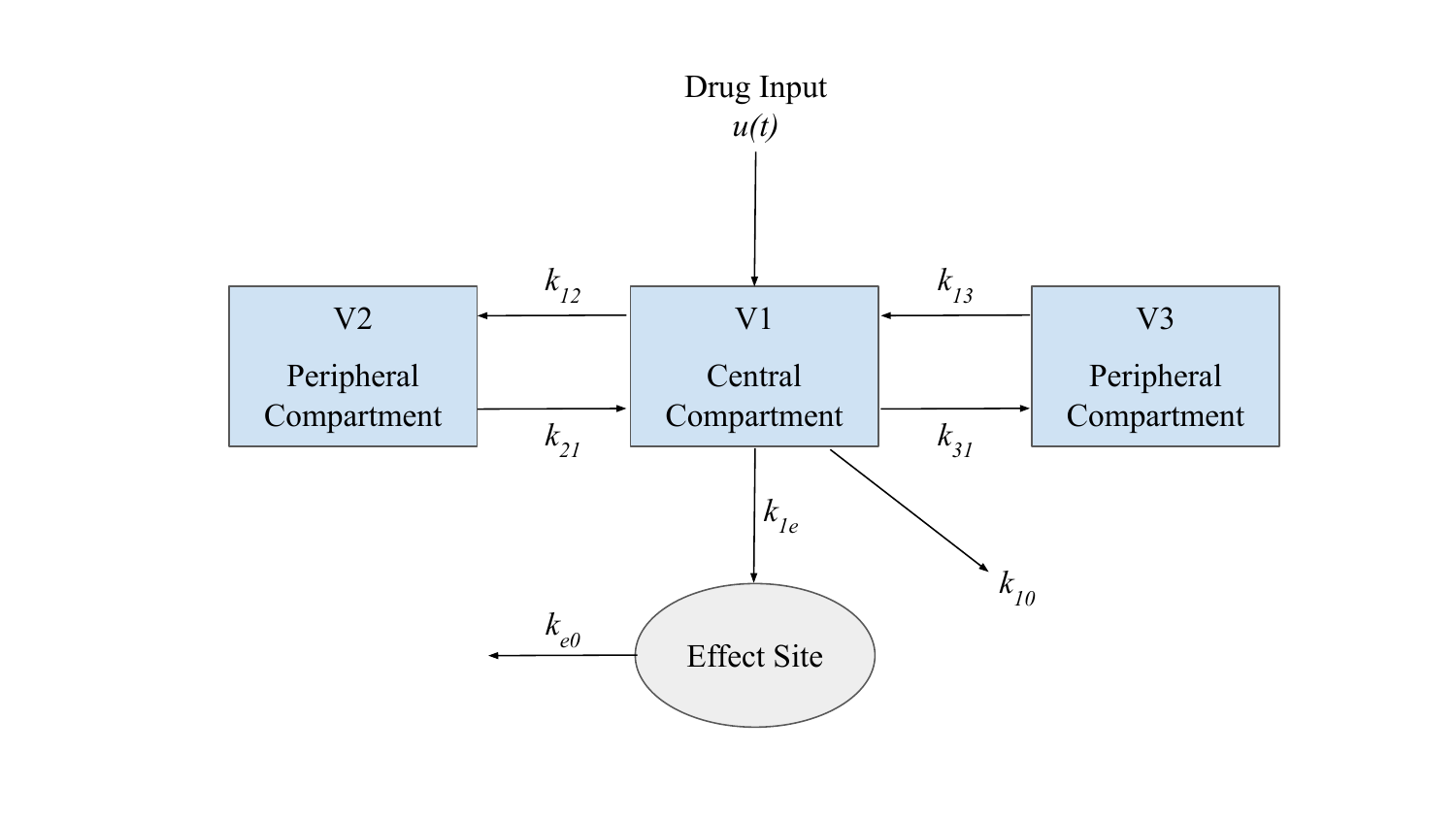}
    \caption[Three Compartments Pharmacokinetics Model]{Three Compartments Pharmacokinetics Model.}
    \label{fig:3comp-model}
\end{figure*}

\begin{figure*}
    \centering
    \includegraphics[width=\linewidth]{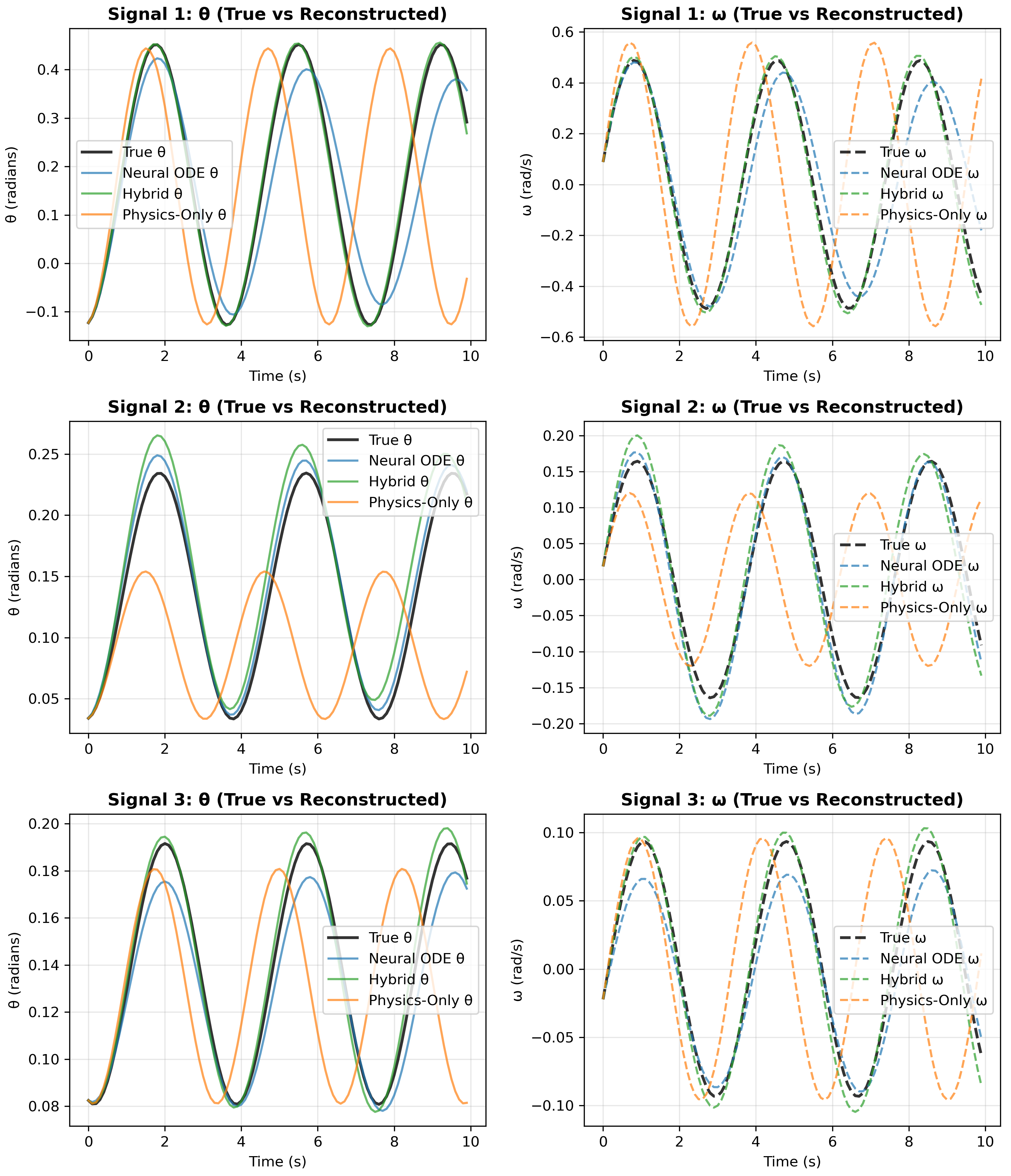}
    \caption[Prediction examples of pendulum dynamics]{Prediction examples of pendulum dynamics.}
    \label{fig:example_pendulum_dynamics}
\end{figure*}

\begin{figure*}
    \centering
    \includegraphics[width=\linewidth]{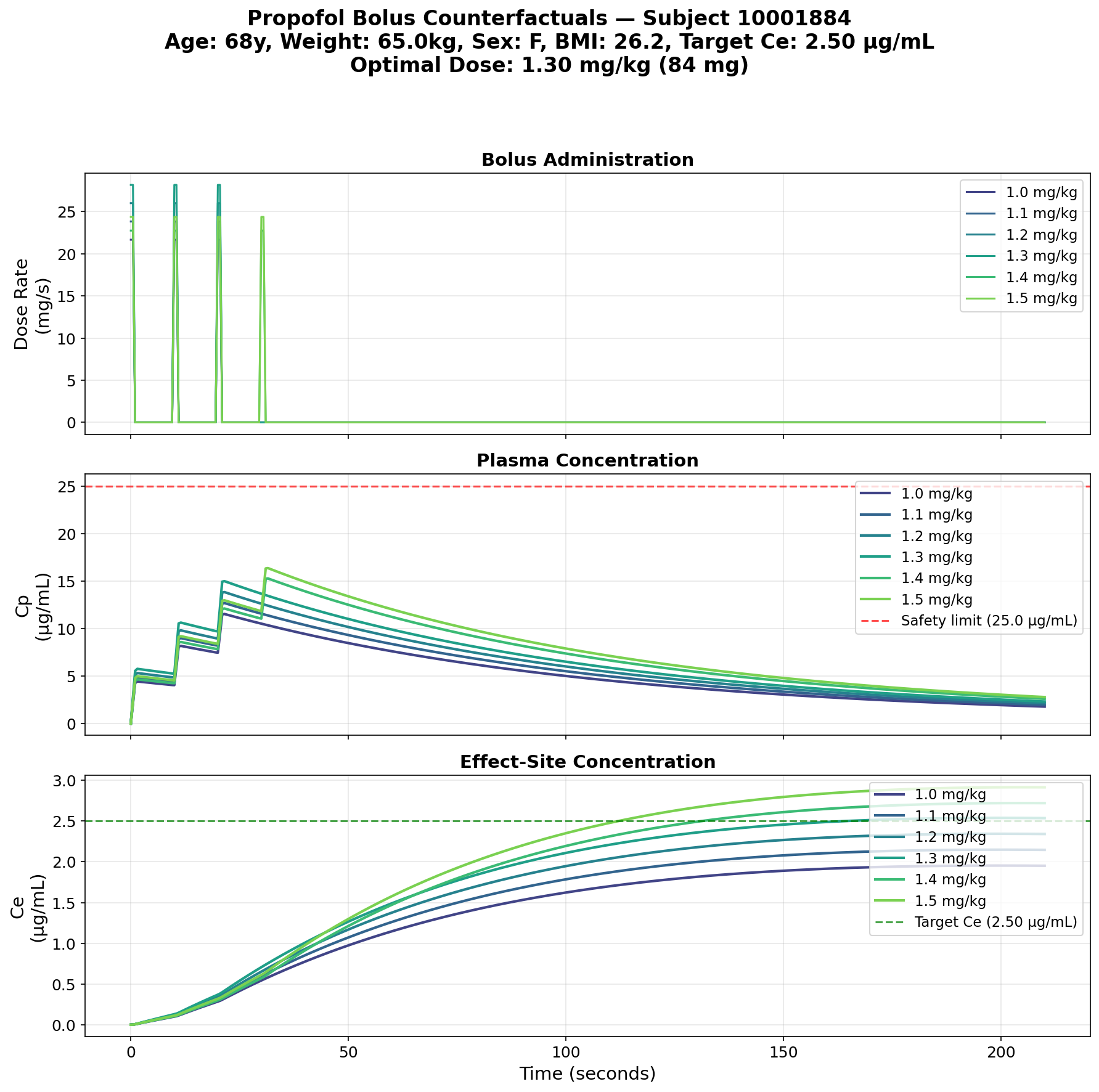}
    \caption[Optimal Dose Selection Simulation]{Optimal Dose Selection Simulation.}
    \label{fig:patient_counterfactuals_example}
\end{figure*}

\begin{figure*}
    \centering
    \includegraphics[width=\linewidth]{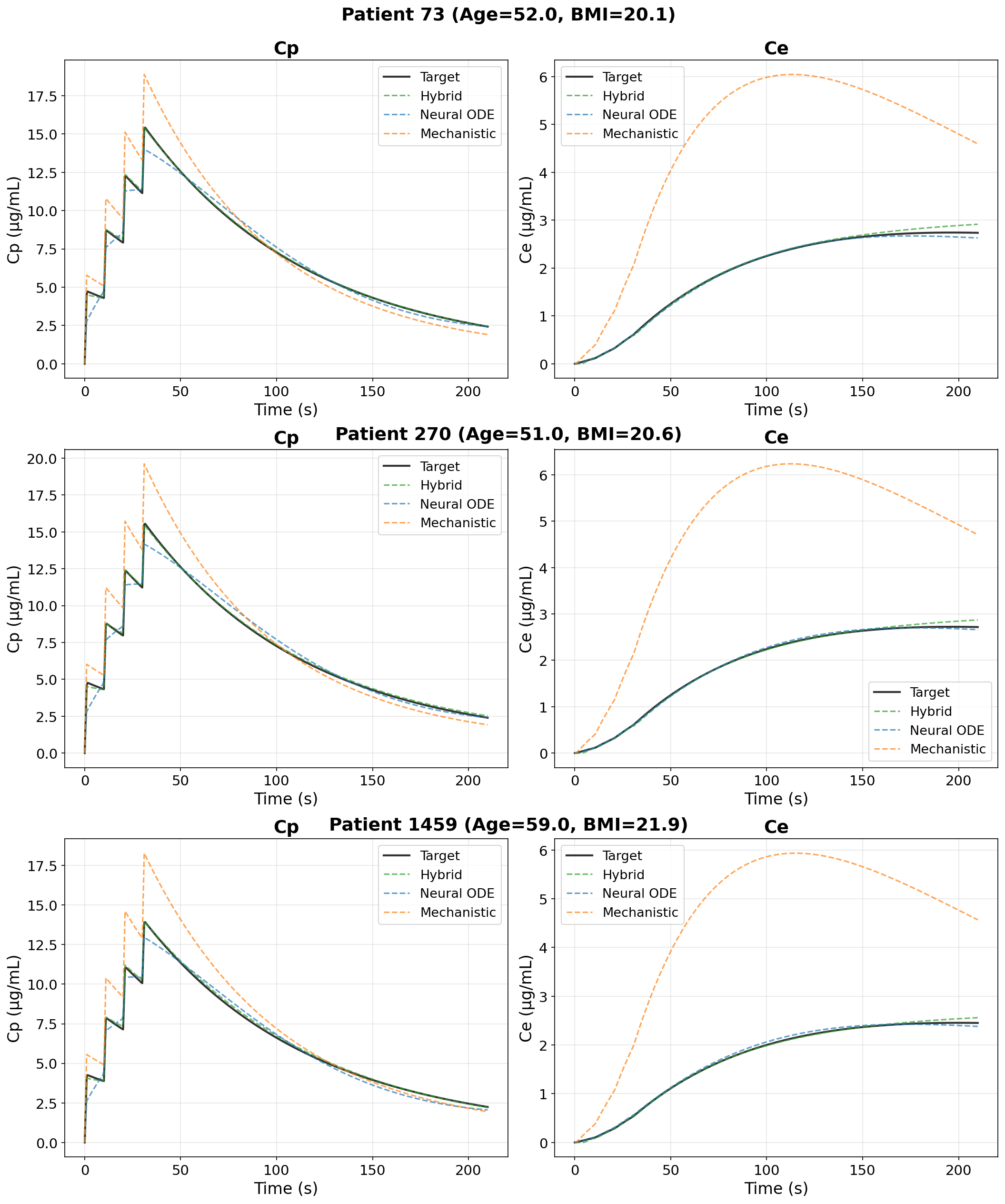}
    \caption[Prediction example of propofol pharmacokinetics dynamics]{Prediction example of propofol pharmacokinetics dynamics.}
    \label{fig:example_propofol_dynamics}
\end{figure*}

\end{document}